\documentclass[10pt,twocolumn,letterpaper]{article}

\usepackage{iccv}
\usepackage{times}
\usepackage{epsfig}
\usepackage{graphicx}
\usepackage{amsmath}
\usepackage[accsupp]{axessibility} 

\usepackage[pagebackref=true,breaklinks=true,letterpaper=true,colorlinks,bookmarks=false]{hyperref}

\usepackage{amsfonts}
\usepackage{array}
\usepackage{booktabs}
\usepackage{color}
\usepackage{enumitem}
\usepackage{float}
\usepackage{makecell}
\usepackage{mathtools}
\usepackage{multirow}
\usepackage{setspace}
\usepackage{subcaption}
\usepackage{threeparttable}
\usepackage{xfrac}

\newcolumntype{L}[1]{>{\raggedright\let\newline\\\arraybackslash\hspace{0pt}}m{#1}}
\newcolumntype{C}[1]{>{\centering\let\newline\\\arraybackslash\hspace{0pt}}m{#1}}
\newcolumntype{R}[1]{>{\raggedleft\let\newline\\\arraybackslash\hspace{0pt}}m{#1}}

\setlist[itemize]{noitemsep, topsep=0pt}
\setlist[enumerate]{noitemsep, topsep=0pt}

\newcommand{\Sota}{State-of-the-art}

\newcommand{\parens}[1]{\left(#1\right)}
\newcommand{\braces}[1]{\left\{#1\right\}}
\newcommand{\bracks}[1]{\left[#1\right]}

\iccvfinalcopy 

\ificcvfinal\fi

\begin{document}

\title{Show Me What I Like: Detecting User-Specific Video Highlights Using Content-Based Multi-Head Attention}

\author{
Uttaran Bhattacharya\thanks{Work partially done while Uttaran was an intern at Adobe Research}\\
University of Maryland\\
College Park, MD, USA\\
\and
Gang Wu\\
Adobe Research\\
San Jose, CA, USA\\
\and
Stefano Petrangeli\\
Adobe Research\\
San Jose, CA, USA\\
\and
Viswanathan Swaminathan\\
Adobe Research\\
San Jose, CA, USA\\
\and
Dinesh Manocha\\
University of Maryland\\
College Park, MD, USA\\
}

\maketitle
\ificcvfinal\thispagestyle{empty}\fi

\begin{abstract}
    We propose a method to detect individualized highlights for users on given target videos based on their preferred highlight clips marked on previous videos they have watched. Our method explicitly leverages the contents of both the preferred clips and the target videos using pre-trained features for the objects and the human activities. We design a multi-head attention mechanism to adaptively weigh the preferred clips based on their object- and human-activity-based contents, and fuse them using these weights into a single feature representation for each user. We compute similarities between these per-user feature representations and the per-frame features computed from the desired target videos to estimate the user-specific highlight clips from the target videos. We test our method on a large-scale highlight detection dataset containing the annotated highlights of individual users. Compared to current baselines, we observe an absolute improvement of 2--4\% in the mean average precision of the detected highlights. We also perform extensive ablation experiments on the number of preferred highlight clips associated with each user as well as on the object- and human-activity-based feature representations to validate that our method is indeed both content-based and user-specific.
\end{abstract}

\maketitle

\begin{figure*}[t]
    \centering
    \includegraphics[width=\textwidth]{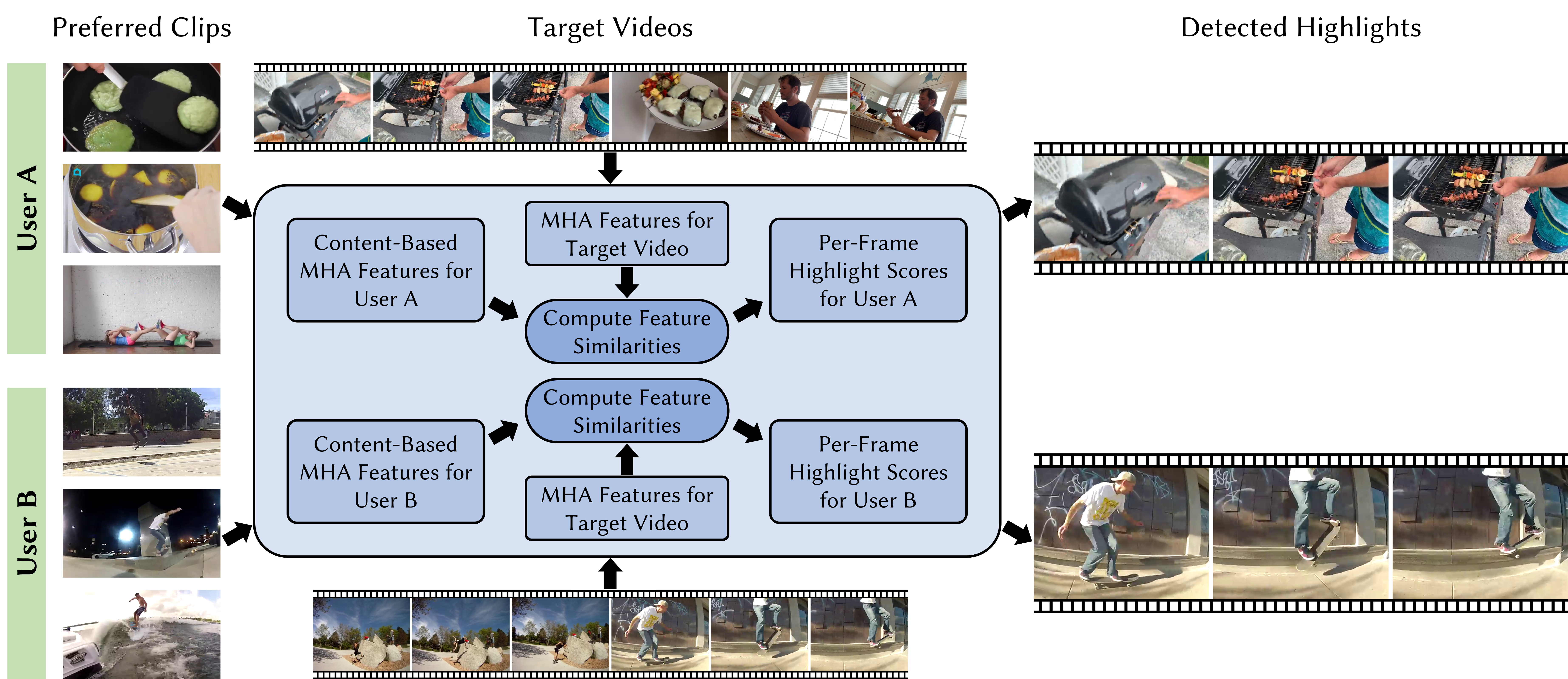}
    \caption{\textbf{User-Specific Highlights for a Variety of Users and Target Videos.} For each user, we consider a set of highlight clips denoting their individual preferences (left) and detect highlights for them (right) on different target videos (center top and bottom). Given the users' overall highlight preferences, our method employs a multi-head attention (MHA) mechanism to learn which segments of the target videos are relevant highlights based on feature similarities between them (center block). For example, our method learns that user A prefers watching cooking and workout videos. Therefore, given a target video containing cooking and eating, our method identifies that only the cooking segments are relevant between the preferred clips and the target videos, and detects those as highlights. Similarly, our method learns that user B prefers skating and surfing videos. Therefore, given a video containing parkour and skating activities, our method detects only the skating segments as highlights. Overall, our method significantly advances the state-of-the-art in user-specific highlight detection given a diverse, large-scale dataset of user preferences and target videos.
    }
    \label{fig:teaser}
\end{figure*}

\section{Introduction}\label{sec:intro}
The rapid growth in the number of unedited videos due to the commercial availability of cameras and video sharing platforms requires the development of automated tools for condensing videos to their most interesting or highlightable moments. This problem is commonly referred to as highlight detection~\cite{lsvm_dsh}. It makes indexing and browsing videos manageable and expedites large-scale previewing, sharing, and recommending of videos~\cite{adaptive_fcsn}. Ongoing research efforts have led to the development of efficient content-based highlight detection methods~\cite{lsvm_dsh,phd2,fcsn,highlightme}. More recently, highlight detection methods have also considered viewers' preferences to provide individualized or \textit{user-specific} highlights~\cite{adaptive_fcsn}, making these methods more practically relevant. However, current user-specific methods either require the expensive computation of shot boundaries to divide videos into highlightable and non-highlightable segments~\cite{phd2} or uniformly pool the users' selected highlights to compute their highlight preferences~\cite{adaptive_fcsn}.

In practice, users' preferences may not be uniformly distributed across their preferred highlight clips but vary significantly based on the clip contents, especially in relation to the target videos. For example, in Fig.~\ref{fig:teaser}, the cooking clips are more relevant indicators of user A's preferences compared to the workout clips, given the target video containing cooking and eating. Thus, combining the users' preferred clips to learn their relevant highlight preferences for each target video requires learning the relevant features describing the highlightable content in each preferred clip. Subsequently, these relevant features need to be mapped to the contents of the target videos where the highlights are to be detected, requiring an automatic strategy to estimate the segments of the target videos that are similar to the relevant features from the preferred clips and, therefore, highlightable. Thus, for user A, a user-specific highlight detection approach needs to detect that the cooking segments in the target video are relevant to the user's preferred cooking clips, and that both the user's preferred workout clips and the eating segments in the target video are irrelevant in the current context.

Detecting highlights based on the similarities between the preferred clips and the target videos can also be expressed as an attention-based retrieval problem. An attention mechanism adaptively weighs the keys of different key-value pairs based on their relative importance to a given query to predict the most suitable responses to the query~\cite{transformer}. Depending on the data paradigm of the key, the value, and the query, attention mechanisms are used in a wide variety of tasks, including tasks in natural language understanding~\cite{bert}, text-based image and video retrieval~\cite{image_retrieval}, object and action recognition in images and videos~\cite{obj_rec_attn,action_rec_attn}, and visual question answering~\cite{vqa}. In the case of user-specific highlight detection, the key, value, and query need to be based on the video contents, \textit{i.e.}, follow the paradigm of content-based highlight detection~\cite{lsvm_dsh,fcsn,highlightme} to perform meaningful retrieval of the highlightable clips per user.

\paragraph{Main Contributions.}
In this paper, we consider the visual content of the videos for highlight detection. Specifically, for both the preferred clips and the target videos, we consider the constituent non-human entities commonly clubbed as ``objects''~\cite{yolov5}, and the human activities expressed with their pose movements. We follow the current paradigm of leveraging the presence of and interactions between human activities and non-human entities, either directly or indirectly, for content-based highlight detection~\cite{lsvm_dsh,phd2,highlightme}.
We design an attention mechanism to leverage both the objects and the pose-based activities in the users' preferred highlight clips to detect user-specific highlights in different target videos. We only consider the preferred highlight clips and not the corresponding full videos, \textit{i.e.}, we do not require information about clips the users did \textit{not} select as highlights. Each highlight clip marked by the users contains partial information about their highlight preferences, which we store in latent features learned from the objects and the poses detected in those clips. We pool these object- and pose-based features using learned weights to obtain combined features representing the users' overall highlight preferences. Given target videos for each user, we then query the object- and the pose-based features of the target videos against the combined features of the corresponding user to detect highlight clips for them in the target videos. To summarize, our main contributions are the following:
\begin{itemize}
    \item \textbf{User-Specific Highlight Detection.} We leverage the video contents, in terms of the objects and the pose-based human activities, between the users' preferred highlight clips and the target videos to detect the user-specific highlights in the target videos.
    \item \textbf{Multi-Head Attention Mechanism.} We design an multi-head attention mechanism that learns content-based features per user from the users' preferred highlight clips. We also learn content-based features from the target videos and query those against the per-user features to detect the user-specific highlights.
    \item \textbf{\Sota~performance.} We perform extensive experiments on the benchmark personal highlight detection dataset (PHD$^2$)~\cite{phd2}, containing more than 6,000 testing videos, to show that our method improves the mean average precision of highlight detection by an absolute 2--4\% over the current baselines, as well as the F-score in the related problem of video summarization by an absolute 4--12\% over the current baselines on the benchmark SumMe dataset~\cite{summe}. Further, we show the qualitative performance of our method as full videos in our supplementary material.
\end{itemize}

\section{Related Work}\label{sec:rw}
We briefly review the prior work in highlight detection and the related problem of video summarization. For a more extensive discourse, including the development of multimedia-based techniques for content-based clustering, scene understanding, and temporal variance optimization, we refer the readers to the works of~\cite{video_clustering,graph_modeling_video_summ,video_abstraction_survey,looking_at_viewer}. Since our goal is to learn user-specific highlights from their past highlight preferences, our problem is also related to collaborative filtering, and we review prior work there as well.

\paragraph{Highlight Detection.}
Highlight detection aims to detect the most interesting moments in videos~\cite{video_abstraction_survey,lsvm_dsh}. Since ``interesting'' is subjective for the viewer, highlight detection methods rely on the availability of annotated highlights to learn the interesting moments. Depending on the nature of the annotations, we can broadly classify these methods into fully supervised, weakly supervised, or unsupervised approaches. Fully supervised approaches assume that, for each video, highlight and non-highlight clips are available, and typically optimize variants of a pairwise ranking loss that ranks the highlight clips higher than the non-highlight clips~\cite{lsvm_dsh,video2gif,pairwise_deep_ranking,attn_deep_ranking,deep_ranking_360_video,region_based_deep_ranking,s2n}. Weakly supervised and unsupervised approaches do not require highlight annotations but instead utilize exemplars or video metadata. Exemplar-based methods use open-source images depicting highlightable moments, which guide the ranking of the individual shots or frames in the videos~\cite{triplet_deep_ranking}, \textit{e.g.}, a surfing image used to detect highlights in a surfing video. Video metadata include information on the video domain~\cite{rrae}, \textit{e.g.}, collecting all edited videos with ``surfing'' in the title to learn surfing highlights, using the video duration itself as a weak signal with the argument that shorter videos are more likely to be edited and therefore have a higher percentage of highlightable moments~\cite{less_is_more}, and even focusing on certain subjects in the videos, such as explicitly leveraging human activities to learn highlights in human-centric videos~\cite{highlightme}. While weakly supervised and unsupervised methods alleviate over-reliance on annotated data, they are designed to be content-based and cannot adapt to the individual preferences of different users.

\paragraph{Preference-Based Highlight Detection.} These methods solve the more challenging problem of learning individualized highlights for the users based on their annotated highlight preferences. Existing methods solve this problem by learning to score highlight frames higher than other frames conditioned on the users' preferences~\cite{phd2}, learning parameters based on the users' preferred clips to guide a content-based highlight detection network~\cite{adaptive_fcsn}, or learning the users' overall preferences by averaging over features learned from their preferred clips~\cite{prnet}. Other methods do not depend on the video contents but rank the clips in a video for each user based on the durations of their selected clips using a recommendation algorithm~\cite{personalized_video_summ}. Our work solves the same problem of predicting preference-based user-specific highlights and improves on the performances of these methods.

\paragraph{Video Summarization.}
Video summarization aims to condense videos into the summaries of their contents. Summaries are typically presented in the form of storyline graphs~\cite{storyline_graph_1,storyline_graph_2}, keyframe sequences~\cite{egocentric_video_summ}, clips~\cite{summe,retro_video_summ}, as well as a mixture of formats~\cite{rule_them_all}. Video summarization can also be broadly classified into fully supervised, weakly supervised and unsupervised approaches. Fully supervised approaches leverage annotated summaries, and learn relevant video subsets for summaries based on their relative importance in the videos~\cite{diverse_sequential_subset,egocentric_video_summ,summe,submodular_mixtures_1,submodular_mixtures_2,sup_video_summ,retro_video_summ,hierarchical_rnn_summ,attn_summ}. Weakly supervised approaches rely on learning the relevant contents for a summary based on exemplars such as images and edited videos~\cite{storyline_graph_1,web_image_prior_summ,weakly_sup_summ,unpaired_data_video_summ}, or leveraging keywords in the video title to learn the relevant content-based features for summaries~\cite{quasi_real_time_summ,tvsum,category_specific_summ,collab_summ}. Unsupervised approaches learn relevant information directly from the video contents, such as scenes that commonly co-occur in videos~\cite{mbf}, and temporal consistency across frames and shots to understand summary boundaries~\cite{joint_summ,unsup_video_summ,fcsn,retro_video_summ}. Other approaches for video summarization define and estimate specific parameters determining summaries, such as coherence~\cite{story_driven_summ}, diversity, and representativeness~\cite{collab_summ,drl_summ}. Yet other approaches consider user preferences in the form of the text queries they use to search videos~\cite{text_query_based_prefs_1,text_query_based_prefs_2,text_query_based_prefs_3,text_query_based_prefs_4}, their feedback on individual summaries~\cite{individual_summaries}, and their preferences between pairs of summaries~\cite{pairwise_summaries}, to learn user-specific summaries. Highlight detection methods learn to leverage video contents using techniques similar to those employed for video summarization, albeit with different training objectives. Further, designing highlight detection methods using user-specific feedback requires the methods to be interactive at inference time and relies on the users' availability and ability to provide feedback. This is complementary to learning user-specific highlights based only on the preferred highlight clips, which streamlines the inference, and the two approaches can be combined depending on the use cases.

\paragraph{Collaborative Filtering.}
Collaborative filtering aims to predict novel user preferences based on their preference history as well as the preferences of other users~\cite{collaborative_filtering}. It is widely used in tackling recommendation problems, such as the Netflix challenge~\cite{netflix_challenge} and online video recommendations~\cite{youtube_recommendations}. However, extending collaborative filtering to highlight detection requires the availability of multiple user responses to each clip within each video in a large training set of videos. Making such an approach practically feasible firstly requires the availability of multiple user responses for the each clip, which is not the case in current highlight detection datasets~\cite{phd2}. Secondly, it requires us to define clip boundaries apriori; otherwise, each video can contain infinitely many clips. This, in turn, severely limits what the users can choose as their highlight preferences, as they can only detect highlight clips of fixed lengths. For these limitations, collaborative filtering is not amenable to user-specific highlight detection. By contrast, our highlight detection method works at the frame level and uses variable-length preferred clips marked by the users to detect their individualized highlights.

\section{User-Specific Highlight Detection}

\begin{figure*}[t]
    \centering
    \includegraphics[width=\textwidth]{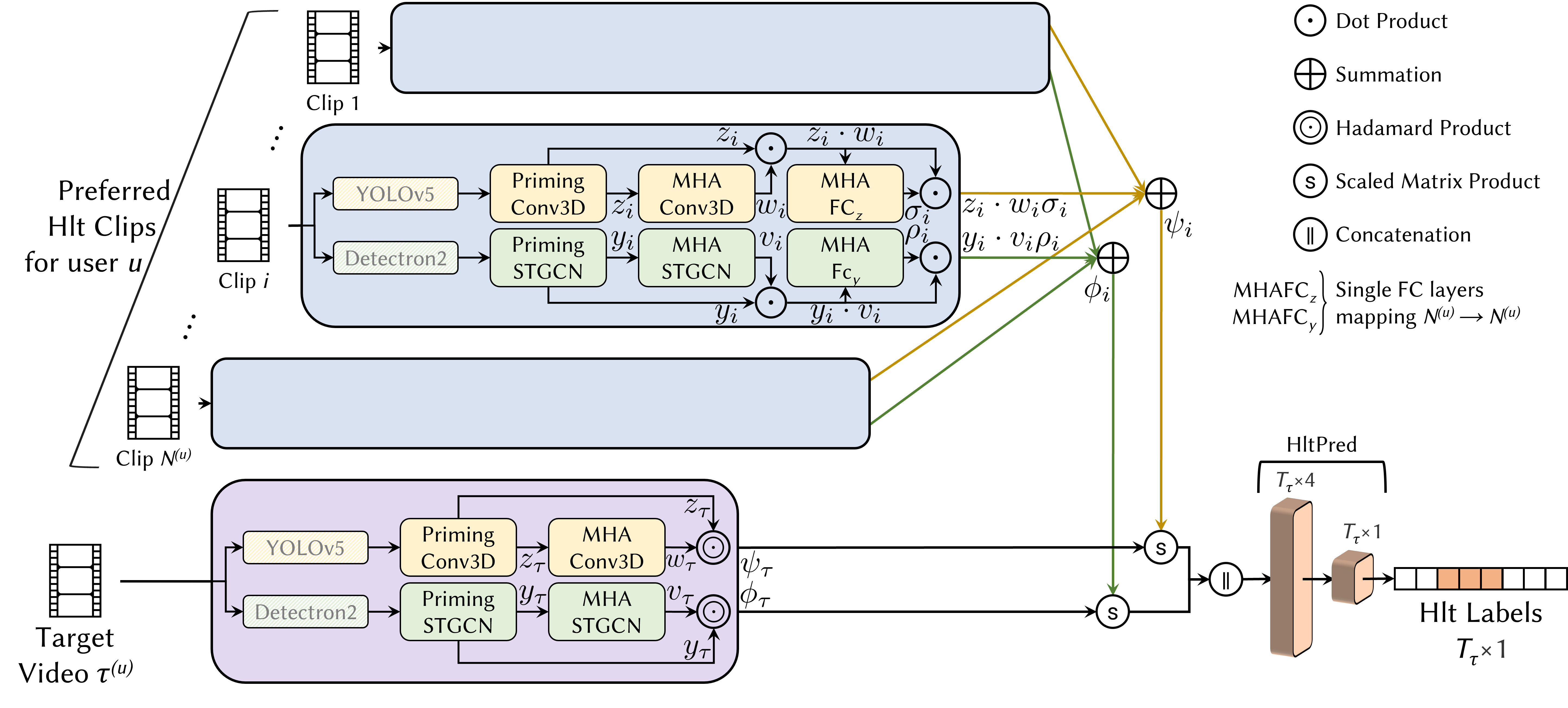}
    \caption{\textbf{Our User-Specific Highlight Detection Network.} For each preferred clip $i$, we use the two priming blocks to map the object-based features and the pose-based features to respective features $z_i$ and $y_i$. We use these features to learn the per-frame weights $w_i$ and $v_i$ using multi-head attention (MHA), perform per-frame attention pooling, learn the per-clip weights $\sigma_i$ and $\rho_i$ using MHA again, and fuse the per-clip features using weighted summation to get the fused features $\psi_i$ and $\phi_i$. For each target video $\tau$, we train a separate set of attention priming and MHA layers to obtain fused features $\psi_\tau$ and $\phi_\tau$. We compute the similarities between the fused features of the preferred clips and the target video using scaled matrix products and concatenate and map the resultant features to per-frame highlight scores for the target video using a fully-connected prediction block.}
    \label{fig:ush_network}
\end{figure*}

\begin{figure}[t]
    \centering
    \includegraphics[width=\columnwidth]{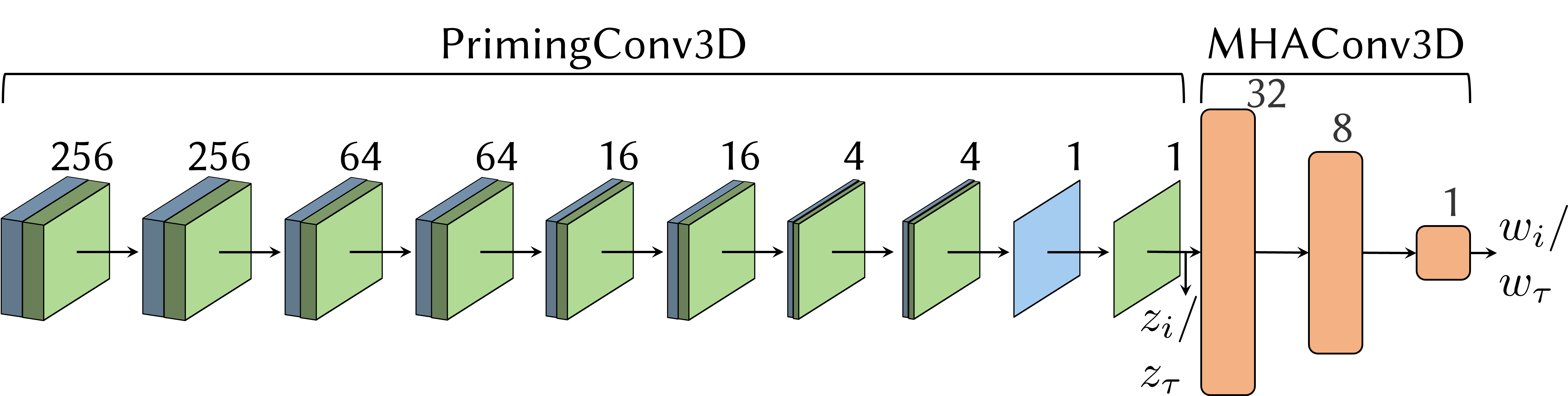}
    \caption{\textbf{Priming and MHA for Objects.} Priming on the YOLOv5 ~\cite{yolov5} features using 3D convolutions (blue blocks) and 3D batch norms (green blocks). We use the attention-primed features to learn the per-frame attention weights $w_i$ and $w_\tau$ using fully-connected layers (orange blocks).}
    \label{fig:conv3d_priming_and_attn}
\end{figure}

Given preferred highlight clips marked by users and different target videos, our goal is to detect highlight clips per user from the target videos that are similar in content to the user's preferred highlight clips. For each user $u$, we consider $P^{\parens{u}} = \braces{i\,|\, i = 1, \dots, N^{\parens{u}}}$ preferred highlight clips, and a target video $\tau^{\parens{u}}$. For numerical consistency, we consider each highlight clip to be $T$ frames long and zero-pad all videos shorter than $T$ frames. In the subsequent text, we assume all the variables are for user $u$ unless stated otherwise and drop the superscript $\parens{u}$ for brevity of notation.

For each user, our objective is to predict the highlight score $s_j \in \bracks{0, 1}$ for each frame $j = 1, \dots, T_\tau$ for the target video, where $T_\tau$ is the number of frames in the target video (we zero-pad all videos shorter than $T_\tau$ frames). The highlight score $s_j$ determines how highlightable a frame is, with $1$ being the highest score. We require our highlight scores to be relative to the contents of both the target video and the user's preferred highlight clips. Thus, we aim to learn a scoring function $\mathcal{S}$ such that
\begin{equation}
    \braces{s_1, \dots, s_{T_\tau}} = \mathcal{S}\parens{\tau, P}. \label{eq:highlight_scores}
\end{equation}

\subsection{Attention Priming}
We consider the contents to be both the non-human entities or ``objects''~\cite{yolov5} and the pose-based human activities, and explicitly leverage their presence in all the clips. For each frame of each clip, we obtain pre-trained features $x_{\textrm{YOLOv5}}$ from the penultimate layer of the YOLOv5 network~\cite{yolov5}, which contains information on the object categories and their spatial positions. We also obtain pre-trained features $x_{\textrm{Detectron2}}$ from the penultimate layer of the Detectron2 network~\cite{detectron2} containing information on the human poses. Our attention priming transforms these pre-trained object and pose features into lower-dimensional features for learning the per-frame attention weights for multi-head attention fusion. This provides the twin benefits of leveraging the localized information common to all the pre-trained features using the same set of convolutional layers, and reducing the parameter load required to learn the actual attention weights. It also ensures our network avoids memorization of the attentions learned for the pre-trained features.

\begin{figure}[t]
    \centering
    \includegraphics[width=\columnwidth]{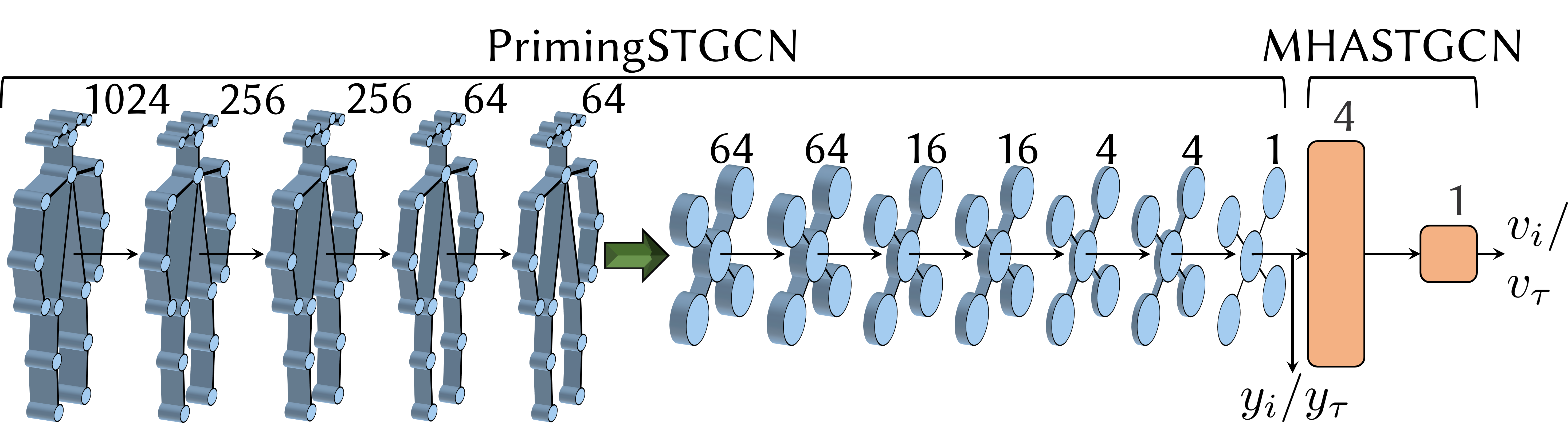}
    \caption{\textbf{Priming and MHA for Poses.} Priming on the Detectron2~\cite{detectron2} features using spatial temporal graph convolutions with feature pooling (green arrow) on the five kinematic chains: trunk, two arms and two legs. We use the attention-primed features to learn the per-frame attention weights $v_i$ and $v_\tau$ using fully-connected layers (orange blocks).}
    \label{fig:stgcn_priming_and_attn}
\end{figure}

To perform attention priming, we pass the YOLOv5 $x_{\textrm{YOLOv5}}$ and the Detectron2 features $_{\textrm{Detectron2}}$ through separate networks to get $z \in \mathbb{R}^{T \times d_z}$ and $y \in \mathbb{R}^{T \times d_y}$ respectively, which we use to learn attention weights. For priming the YOLOv5 features, we use a series of 3D convolutions, denoted $\textrm{PrimingConv3D}$ (Fig.~\ref{fig:conv3d_priming_and_attn}), to leverage the spatial and temporal adjacency of the features $x_{\textrm{YOLOv5}}$, as
\begin{equation}
    z = \textrm{PrimingConv3D}\parens{x_{\textrm{YOLOv5}}; \theta_{\textrm{PrimingConv3D}}},
\end{equation}
where $\theta_{\textrm{PrimingConv3D}}$ are trainable parameters. Similarly, for priming the Detectron2 features, we use a series of spatial temporal graph convolutions, denoted $\textrm{PrimingSTGCN}$ (Fig.~\ref{fig:stgcn_priming_and_attn}), to leverage adjacencies between the body joints that comprise the features $x_{\textrm{Detectron2}}$, as
\begin{equation}
    y = \textrm{PrimingSTGCN}\parens{x_{\textrm{Detectron2}}; \theta_{\textrm{PrimingSTGCN}}},
\end{equation}
where $\theta_{\textrm{PrimingSTGCN}}$ are trainable parameters.

\subsection{Multi-Head Attention and Fusion}
For each preferred highlight clip $i$, we denote the attention-primed features as $z_i \in \mathbb{R}^{T \times d_z}$ and $y_i \in \mathbb{R}^{T \times d_y}$. Each $\parens{z_i, y_i}$ contains information on the user's highlight preferences in clip $i$, which is different from the user's highlight preferences in all the other preferred clips. Thus, to obtain the user's overall highlight preferences, we consider a multi-head attention (MHA) mechanism that learns relative weights at both the frame and the clip levels.

At the frame level, the number of attention heads is the number of frames $T$ for the preferred clips and $T_\tau$ for the target video. The parameters determining the attention weights are shared for all the preferred clips, implying that the order in which our network processes them is irrelevant. This is the desired behavior since the preferred clips need not have any content overlap or any specified ordering. However, we train a separate set of parameters to learn the attention weights from the target videos since the target video contains both highlightable and non-highlightable segments.

At the clip level, the number of attention heads is the number of preferred clips $N$ that we consider for each user. We use the clip level attention weights to additively fuse the corresponding attention-primed features and then compute the similarity of the fused features to the attention-primed features of the target video. During training, the frame level and the clip level attention weights collectively determine the features describing the relevant contents of the preferred clips given the target videos per user. Conversely, during inference, our network uses these per-user relevant features to detect the highlightable frames from the target videos.

For our MHA mechanism, we train a pair of attention heads $\textrm{MHAConv3D}$ (Fig.~\ref{fig:conv3d_priming_and_attn}) and $\textrm{MHASTGCN}$ (Fig.~\ref{fig:stgcn_priming_and_attn}) to produce frame level attention weights $w_i \in \mathbb{R}^T$ and $v_i \in \mathbb{R}^T$ as
\begin{align}
    w_i &= \textrm{MHAConv3D}\parens{z_i; \theta_{\textrm{MHAConv3D}}}, \\
    v_i &= \textrm{MHASTGCN}\parens{y_i; \theta_{\textrm{MHASTGCN}}},
\end{align}
where $\theta_{\textrm{MHAConv3D}}$ and $\theta_{\textrm{MHASTGCN}}$ are trainable parameters. We train another pair of attention heads $\textrm{MHAFC}_z$ and $\textrm{MHAFC}_y$ (Fig.~\ref{fig:ush_network}, rightmost layers in central block) to get clip level attention weights $\sigma_i \in \mathbb{R}$ and $\rho_i \in \mathbb{R}$ from the weighted frame level features as
\begin{align}
    \sigma_i &= \textrm{MHAFC}_z\parens{z_i^\top w_i; \theta_{\textrm{MHAFC}_z}}, \\
    \rho_i &= \textrm{MHAFC}_y\parens{y_i^\top v_i; \theta_{\textrm{MHAFC}_y}},
\end{align}
where $\theta_{\textrm{MHAFC}_z}$ and $\theta_{\textrm{MHAFC}_y}$ are trainable parameters. We obtain the fused, attention-weighted features $\psi \in \mathbb{R}^{d_z}$ and $\phi \in \mathbb{R}^{d_y}$ as
\begin{align}
    \psi &= \sum_{i = 1}^N z_i^\top w_i \sigma_i, \\
    \phi &= \sum_{i = 1}^N y_i^\top v_i \rho_i.
\end{align}

Parallelly, we compute the per-frame attention-weighted features of the target video as $\psi_\tau = z_\tau \bigodot w_\tau \in \mathbb{R}^{T_\tau \times d_z}$ and $\phi_\tau = y_\tau \bigodot v_\tau \in \mathbb{R}^{T_\tau \times d_y}$. Given $\psi_\tau$, $\phi_\tau$, $\psi$, and $\phi$, we use scaled matrix products to compute features $h_z \in \mathbb{R}^{T_\tau}$ and $h_y \in \mathbb{R}^{T_\tau}$ representing their pairwise similarities, as
\begin{align}
    h_z &= \textrm{softmax}\parens{\psi d_z^{\sfrac{-1}{2}}} \cdot \psi_\tau \label{eq:combined_features_obj}, \\
    h_y &= \textrm{softmax}\parens{\phi d_y^{\sfrac{-1}{2}}} \cdot \phi_\tau. \label{eq:combine_features_pose}
\end{align}

We then use a predictor block $\textrm{HltPred}$ (Fig.~\ref{fig:ush_network}, bottom-right block) that fuses the features $h_z$ and $h_y$ using concatenation, and reduces the fused features to per-frame highlight scores $s \in \bracks{0, 1}^{T_\tau}$, as
\begin{equation}
    s = \textrm{HltPred}\parens{h_z \parallel h_y; \theta_{\textrm{HltPred}}},
\end{equation}
where $\theta_{\textrm{HltPred}}$ are trainable parameters.

\section{Training and Inference}
We discuss the loss function we use to train our network, the implementation details, and how we use our network to detect user-specific highlights on test videos.

\subsection{Loss Function}\label{subsec:loss_functions}
Our network predicts a highlight score between $0$ and $1$ for each frame in the target video. For training, we consider a frame to be a highlight if and only if its score is above a predetermined threshold $\zeta$. We train our network using a combination of a label loss, a margin loss, and a sparsity loss, assuming a threshold $\zeta=0.5$.

\begin{itemize}
    \item \textbf{Label loss} $L_{\tau j}$ for each frame $j$ in target video $\tau$ measures the weighted cross entropy loss between the ground-truth highlight labels $y_{\tau j}$ ($1$ for highlight, $0$ otherwise) and the predicted highlight scores $s_{\tau j}$, as
    \begin{equation}
        L_{\tau j} = -wy_{\tau j}\log\parens{s_{\tau j}} - \parens{1 - y_{\tau j}}\log\parens{1 - s_{\tau j}},
    \end{equation}
    where $w$ denotes the relative weight assigned to the highlight class. The label loss is the baseline loss that guides the training process as a binary classification problem.
    
    \item \textbf{Margin loss} $M_{\tau j}$ for each frame $j$ in target video $\tau$ measures the one-sided distance between the highlight score of that frame and the threshold $\zeta$ as
    \begin{equation}
        M_{\tau j} = \max\parens{0, \widetilde{y}_{\tau j}\parens{\zeta - s_{\tau j}}},
    \end{equation}
    where $\widetilde{y}_{\tau j} = 1$ if $j$ is a highlight frame, and $-1$ otherwise. The margin loss provides additional constraints that the scores for the highlight and the non-highlight frames should be on opposite sides of the threshold $\zeta$.
    
    \item \textbf{Sparsity loss} $S_\tau$ for each target video $\tau$ enforces that the total number of highlight frames should be low, as
    \begin{equation}
        S_\tau = \sum_{j = 1}^T \max\parens{0, \textrm{sign}\parens{s_{\tau j} - \zeta}}.
    \end{equation}
    The sparsity loss incentivizes our network to detect as few highlight frames as possible, following the intuition that highlight frames make up a small fraction of total video~\cite{lsvm_dsh,less_is_more,highlightme}. Moreover, the sparsity loss improves the precision of highlight detection by its design of minimizing the number of frames detected as highlights.
\end{itemize}

Combining the individual loss terms, we write the overall loss function $\mathcal{L}_\tau$ for the target video $\tau$ as
\begin{equation}
    \mathcal{L}_\tau = \frac{1}{T}\sum_{j = 1}^T\parens{L_{\tau j} + M_{\tau j}} + S_\tau.
\end{equation}

Based on our experiments, we did not observe significant performance differences when using relative weights for the margin and the sparsity losses up to five times that of the baseline label loss. Consequently, we propose using the same weights for all the loss terms for simplicity. We also note that none of our losses enforce the temporal continuity of labels, \textit{i.e.}, adjacent frames should have the same labels (highlight or non-highlight) except at the highlight clip boundaries. There are two reasons for this. First, the durations of highlight clips are not fixed. In our experiments, the longest highlight clip can be about seven times the length of the shortest highlight clip for the same video. Thus, defining clip boundaries for enforcing temporal continuity is non-trivial and would, in fact, limit the variety of highlight clips that we can detect during inference. Second, our network explicitly leverages information along both the spatial and the temporal dimensions using 3D convolutions and spatial temporal graph convolutions. Consequently, temporal continuity is implicitly enforced in the learned features and highlight scores, and in practice, we observe the highlight scores varying smoothly across the frames, with only minor variance due to noise.

\begin{table}[t]
    \centering
    \caption{\textbf{Mean Average Precision (mAP) and normalized Meaningful Summary Duration (nMSD) for Highlight Detection.} We report the numbers of all methods on PHD$^2$~\cite{phd2}. \textbf{Bold} indicates best.}
    \label{tab:map_nmsd_phd2}
    \resizebox{\columnwidth}{!}{%
    \begin{threeparttable}
        \begin{tabular}{clrr}
            \toprule
            & Method & mAP & nMSD \\
            \midrule
            \multirow{3}{*}{User-Agnostic} & Random & 0.112 & 0.536 \\
            & FCSN~\cite{fcsn} & 0.152 & - \\
            & HighlightMe~\cite{highlightme} & 0.200 & - \\
            \midrule
            \multirow{6}{*}{User-Specific} & Personalized Summ.~\cite{personalized_video_summ} & 0.216\tnote{*} & 0.288\tnote{*} \\
            & Video2GIF~\cite{video2gif} & 0.158 & 0.420 \\
            & PHD-GIF~\cite{phd2} & 0.166 & 0.402 \\
            & Adaptive-FCSN~\cite{adaptive_fcsn} & 0.168 & - \\
            & PR-Net~\cite{prnet} & 0.187 & - \\
            & \textbf{MHA+Fusion (Ours)} & \textbf{0.228} & \textbf{0.271} \\
            \bottomrule\addlinespace[1ex]
        \end{tabular}
        \begin{tablenotes}\footnotesize
            \item[*] Numbers reported for a subset of the test set consisting of at least 5 selected highlights per user. For a similar test subset, our method has mAP of 0.262 and nMSD of 0.223.
        \end{tablenotes}
    \end{threeparttable}
    }
\end{table}

\subsection{Implementation}
We first extract the YOLOv5~\cite{yolov5} and the Detectron2~\cite{detectron2} features for each frame in both the preferred highlight clips and the target videos in the training dataset. This takes about $30$ seconds per frame on an NVIDIA Tesla A100 GPU. We then train our network using the Adam optimizer~\cite{adam} with a batch size of $36$, an initial learning rate of $1\textsc{e--}4$ that we decay at a rate of $0.999$ per epoch, and a weight decay of $3\textsc{e--}4$. We train for $500$ epochs on $8$ NVIDIA Tesla V100 GPUs at a speed of about $1,480$ seconds per epoch. Once the network is fully trained, inference takes about $9$ seconds for each user using the same GPU configuration. We provide additional details on hyperparameter tuning in our supplementary material.

\subsection{Inference}~\label{subsec:inference}
Similar to the training set up, we consider a set of preferred highlight clips and target videos for each user during inference. We use the preferred highlight clips to learn the features $h_z$ and $h_y$ (Eqs.~\ref{eq:combined_features_obj} and \ref{eq:combine_features_pose}), and subsequently the per-frame highlight scores $s_j$ (Eq.~\ref{eq:highlight_scores}). We also consider the scenario where the users' preferred highlight clips are not available, where our network falls back to content-based highlight detection~\cite{highlightme}. Instead of the softmax operations in Eqs.~\ref{eq:combined_features_obj} and \ref{eq:combine_features_pose}, we perform uniform weighting along the feature dimension, \textit{i.e.}, we have $h_z = \sum_{k = 1}^{d_z} \psi_{\tau k} / d_z$ and $h_y = \sum_{k = 1}^{d_y} \phi_{\tau k} / d_y$.

\begin{table}[t]
    \centering
    \caption{\textbf{F-Score for Video Summarization.} We report the F-scores of all methods on the SumMe dataset~\cite{summe}. \textbf{Bold} indicates best.}
    \label{tab:f_score_summe}
    \begin{tabular}{lr}
        \toprule
        Method & F-Score \\
        \midrule
        SumMe baseline~\cite{summe} & 0.394 \\
        Submodular mixtures~\cite{submodular_mixtures_1} & 0.397 \\
        DPP-LSTM~\cite{sup_video_summ} & 0.386 \\
        Unsup. Adversarial LSTM~\cite{unsup_video_summ} & 0.417 \\
        Sup. Deep RL~\cite{drl_summ} & 0.421 \\
        S$^2$N~\cite{s2n} & 0.433 \\
        Adaptive-FCSN~\cite{adaptive_fcsn} & 0.444 \\
        HighlightMe~\cite{highlightme} & 0.480 \\
        \textbf{MHA+Fusion (Ours)} & \textbf{0.526} \\
        \bottomrule
    \end{tabular}
\end{table}

\section{Experiments and Results}
We provide details on the dataset we use for training and testing our method, the baselines we compare with, and the performance of our method on quantitative metrics. We also show the improvement in performance as we add more preferred highlight clips to our input, the benefits of the two components of object detection and pose detection used in our approach, and the contribution of each of the loss functions we use in training.

\subsection{Dataset}
We evaluate on the personal highlight detection dataset (PHD$^2$) introduced by del Molino and Gygli~\cite{phd2}. This dataset consists of URLs of YouTube videos, IDs of annotators or ``users'', and the segments they selected as highlight clips from those videos as per their preferences. The last video that a user annotated is designated as the target video for that user. Since PHD$^2$ only provides the YouTube URLs, we scraped the videos from YouTube for our experiments, subject to video availability and IP restrictions. Both our training and testing sets consist of up to 15 highlight clips and a target video per user. The highlight clips are between 1 and 672 seconds long and have a mean length of 5.19 seconds. The target videos are between 1 and 37,434 seconds long and have a mean length of 443.29 seconds. The training set contains a total of 6,596 users and 26,390 videos. The testing set contains 727 users and 6,004 videos that do not overlap with any user or video in the training set.

\subsection{Baselines}
We compare with the methods Video2GIF by Gygli et al.~\cite{video2gif}, PHD-GIF by del Molino and Gygli~\cite{phd2}, Personalized Summarization by Panagiotakis et al.~\cite{personalized_video_summ}, Adaptive-FCSN by Rochan et al.~\cite{adaptive_fcsn}, and PR-Net by Chen et al.~\cite{prnet}, which, like our method, consider user-specific history when detecting highlights for target videos. Video2GIF computes shot boundaries in the videos and uses C3D features, used for action recognition, to map the shots to highlight labels and train using a ranking loss. Adaptive-FCSN trains an encoder network to learn affine parameters based on the users' preferred highlight clips and uses those parameters to guide the detection of per-frame highlight labels in the target videos. We also compare with state-of-the-art user-agnostic approaches FCSN by Rochan et al.~\cite{fcsn}, and HighlightMe by Bhattacharya et al.~\cite{highlightme}, which detect highlights based only on the video contents. FCSN utilizes the GoogLeNet backbone to learn image-based features for per-frame highlight detection. HighlightMe leverages pose-based activities and facial expressions in human-centric videos and trains an autoencoder-based architecture to detect per-frame highlights.

\begin{table}[t]
    \centering
    \caption{\textbf{Ablation 1: Changing the Number of Preferred Highlights.} We report the mAPs of all ablated versions on PHD$^2$~\cite{phd2}. \textbf{Bold} indicates best.}
    \label{tab:map_diff_preferred_clips}
    \resizebox{\columnwidth}{!}{%
    \begin{tabular}{lrrrr}
        \toprule
        \# preferred Highlight Clips & 0 & 5 & 10 & \textbf{15 (max)} \\
        \midrule
        mAP & 0.151 & 0.197 & 0.211 & \textbf{0.228} \\
        \bottomrule
    \end{tabular}
    }
\end{table}

\begin{table}[t]
    \centering
    \caption{\textbf{Ablation 2: Using Only One Pre-Trained Backbone for Highlight Detection.} We report the mAPs of all ablated versions on PHD$^2$~\cite{phd2}. \textbf{Bold} indicates best.}
    \label{tab:map_single_component}
    \resizebox{\columnwidth}{!}{%
    \begin{tabular}{lrrr}
        \toprule
        Using backbone & YOLOv5~\cite{yolov5} & Detectron2~\cite{detectron2} & \textbf{both} \\
        \midrule
        mAP & 0.164 & 0.193 & \textbf{0.228} \\
        \bottomrule
    \end{tabular}
    }
\end{table}

\subsection{Quantitative Comparison}
We evaluate the performance of each method using the metrics of mean average precision (mAP) and normalized meaningful summary duration (nMSD). We compute mAP as the mean of the average precision of matching the highlight labels in each target video following~\cite{fcsn,personalized_video_summ,adaptive_fcsn,highlightme,prnet}, and nMSD at a recall rate of 0.5 following~\cite{video2gif,phd2,personalized_video_summ}. As empirical lower bounds on mAP and nMSD, we also report the performance of randomly choosing the highlight frames in the target videos. Table~\ref{tab:map_nmsd_phd2} shows the mAP and the nMSD of each method on PHD$^2$~\cite{phd2}. Our method achieves the best mAP of 0.219 and nMSD of 0.271, outperforming the current best user-agnostic method HighlightMe~\cite{highlightme} by an absolute 2\% and the current best user-specific methods of PR-Net~\cite{prnet} and Personalized Summarization~\cite{personalized_video_summ} by absolute 4\% and 6\% respectively.

Beyond highlight detection, we also evaluate our method on the related problem of video summarization, following the approach of Rochan et al.~\cite{adaptive_fcsn}. Video summarization aims to condense videos into summaries of their contents, which may or may not correspond to the most interesting moments that users annotate as highlights. We test the performance of our trained network on the SumMe~\cite{summe} dataset, which consists of 25 videos totaling about 66 minutes, for video summarization. We compute the F-score of matching the summary frames and compare with the summarization methods listed by Rochan et al.~\cite{adaptive_fcsn}, their own highlight detection method, and the highlight detection method of HighlightMe~\cite{highlightme}. We show the results in Table~\ref{tab:f_score_summe}, where we observe our method improves the F-score by an absolute 4\% over the next-best method of HighlightMe and 12\% over the SumMe baseline. Our results further corroborate Rochan et al.'s argument that methods for video summarization can benefit from using networks with parameters pre-trained using loss functions for highlight detection.

\begin{table}[t]
    \centering
    \caption{\textbf{Ablation 3: Ablating the Training Loss Functions.} We report the mAPs of all ablated versions on PHD$^2$~\cite{phd2}. \textbf{Bold} indicates best.}
    \label{tab:map_ablate_loss}
    \begin{tabular}{lrrrr}
        \toprule
        Excluding loss & label & margin & sparsity & \textbf{none} \\
        \midrule
        mAP & 0.155 & 0.192 & 0.163 & \textbf{0.228} \\
        \bottomrule
    \end{tabular}
\end{table}

\begin{figure*}[t]
    \centering
    \includegraphics[width=\textwidth]{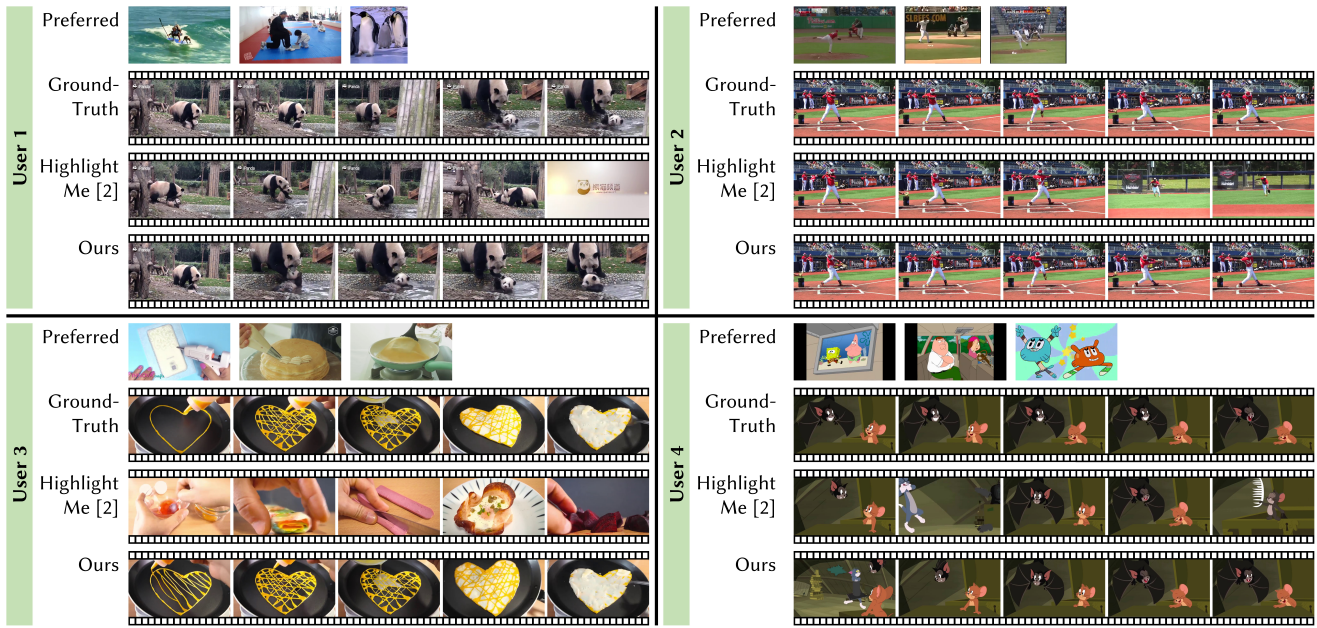}
    \caption{\textbf{Qualitative Results.} We show qualitative results of our method and the current best baseline of HighlightMe~\cite{highlightme} for four users in the testing set of PHD$^2$\cite{phd2}. For each user, we show sample frames of (i) their preferred clips, (ii) ground-truth highlights selected by them from their target videos, (iii) highlights detected by HighlightMe~\cite{highlightme}, and (iv) highlights detected by our method. For each user, we observe that our method matches the ground-truth more closely than HighlightMe~\cite{highlightme}.}
    \label{fig:visual_results}
\end{figure*}

\subsection{Ablation Studies}
We ablate our method in terms of (i) the number of preferred highlight clips used, (ii) the two backbones leveraging objects and poses, and (iii) the loss functions used for training our network.

To evaluate the benefit of using the preferred highlight clips, we train four versions of our network using at most 0, 5, 10, and 15 preferred highlight clips per user. We report the mAPs of the corresponding trained networks on PHD$^2$ in Table~\ref{tab:map_diff_preferred_clips}. We observe a monotonic increase in performance as we increase the number of preferred highlight clips. This shows that our network learns more accurate representations of user-specific highlight preferences as we feed it more data on the preferred highlight clips. At the same time, the rate of increase drops as we increase the number of preferred highlight clips, indicating a diminishing marginal benefit of providing additional clips. We also note that the performance of our method without a substantial number of preferred highlight clips is worse than the user-agnostic method of HighlightMe~\cite{highlightme}. This is because the highlights in PHD$^2$ are largely human-centric, which HighlightMe is fine-tuned to detect. However, the performance of HighlightMe remains fixed, while ours grows and outperforms it as we increase the number of preferred highlight clips. We expect that combining our user-specific approach with the human-centric approach of HighlightMe will lead to the best performance on datasets such as PHD$^2$. For the sake of completeness, we further test the performance of our method without using any preferred clips on a domain-specific highlight dataset (DSH)~\cite{lsvm_dsh} and a title-based summarization dataset (TVSum)~\cite{tvsum}, which focus on content domains such as surfing, cooking, grooming animals, and do not provide user-specific highlight preferences. We report the results in our supplementary material and observe that our method is at par with the current methods performing content-based highlight detection.

We also ablate each of our two network components, the one with the YOLOv5 backbone and the one with the Detectron2 backbone, and train the network using only the remaining component. For these experiments, there is no concatenation of the two components. Instead, we adjust the number of parameters of the first fully-connected layer of the prediction block $\textrm{HltPred}$ (Fig.~\ref{fig:ush_network} bottom-right block) accordingly. We show the mAP of each of these ablated networks on PHD$^2$ in Table~\ref{tab:map_single_component}. We observe a more significant drop in mAP when ablating the component with the Detectron2 backbone, which results in the network not explicitly learning from the human activities in the videos. This is again a consequence of the annotated highlights in PHD$^2$ being largely human-centric.

Lastly, we ablate each of the three loss functions we describe in Sec.~\ref{subsec:loss_functions} and train our network using the remaining loss functions. We report the mAP of these ablated versions in Table~\ref{tab:map_ablate_loss}. Without the label loss, the label assignment becomes random, being only regularized by the margin and the sparsity losses. This results in a sharp drop in the mAP. Without the margin loss, all the predicted highlight scores are closer to the threshold $\zeta$, resulting in more noise in the highlight scores around the threshold $\zeta$ and consequently more confusion between the highlight and the non-highlight labels. Without the sparsity loss, more frames are incorrectly labeled as highlights, again resulting in a sharp drop in the mAP.

\subsection{Qualitative Results}
We show sample visual results in Fig.~\ref{fig:visual_results}. Full video results are available in our supplementary. We show sample frames from the users' preferred highlights, the ground-truth highlights in the users' target videos, and the highlights detected by HighlightMe~\cite{highlightme} and our method. For our method, we observe content similarities between the users' preferred highlights and the target videos, which also largely match the ground-truth. By contrast, HighlightMe~\cite{highlightme} only looks at the target videos and does not adapt to the users' preferences. Our method also fails when the users break their preference patterns in the target videos, or the target videos did not contain any features similar to the preferred highlights. Please refer to our video results for details.

\section{Conclusion}
We have presented a method for user-specific highlight detection based on multi-head attention fusion of features from the users' preferred highlight clips. Our method only requires the highlight clips and not the clips users did not select as highlights, thus significantly reducing the data overhead for training. Our multi-head attention mechanism leverages both objects and human activities in videos to learn highlights based on their spatial and temporal variations and mutual interactions. We also perform experiments to show that our method advances the state-of-the-art in both highlight detection in a large-scale dataset with annotated user preferences and the related problem of video summarization. We also show that, unlike user-agnostic highlight detection methods, the performance of our method increases as we increase the maximum number of preferred highlight clips considered for each user.

\section{Limitations and Future Work}
Our work has some limitations. First, we consider users' highlight preferences only at training time and do not consider user feedback at inference. To this end, we plan to combine our method with complementary methods that rely on user feedback to gradually learn their preferences. Second, our attention mechanism relies on only the non-human entities and the human activities. While this helps us achieve state-of-the-art results currently, we intend to extend to a broader definition of content, including the associated audio, semantic segmentation of the video frames, and other human-centric modalities such as faces. Lastly, we do not fine-tune our method to video categories based on domains (\textit{e.g.}, surfing videos) or the constituent subjects (\textit{e.g.}, human-centric videos). Rather, we design an approach to combine diverse highlight clips assuming the presence of only generic elements, \textit{i.e.}, objects and humans. Nevertheless, our method is suitable for fine-tuning as necessary to improve user-specific highlight detection in various video categories.

\section*{Acknowledgment}
Bhattacharya and Manocha were supported in part by ARO Grants W911NF1910069 and W911NF2110026.

{\small
\bibliographystyle{ieee_fullname}
\bibliography{main}
}

\appendix

\section{Hyperparameter Tuning}\label{sec:hyperparam_tuning}
We have experimented with batch sizes between 2 (minimum possible) and 36 (maximum allowable given our memory constraints), initial learning rates between $1\textsc{e--}5$ (very slow convergence) and $1\textsc{e--}3$ (optimization diverges), learning rate decays between $0.9$ (very slow convergence) and $1$ (optimization oscillates), weight decays between $1\textsc{e--}4$ (optimization oscillates) and $5\textsc{e--}4$ (very slow convergence), and training epochs between $100$ and $1000$ (validation loss saturates around $500$ epochs).

\section{Additional Highlight Detection Results}\label{sec:additional_results}
We show the quantitative performance of our highlight detection method on two additional datasets: Domain-Specific Highlights (DSH)~\cite{lsvm_dsh} for highlight detection and Title-Based Video Summarization (TVSum)~\cite{tvsum} for video summarization. Both these datasets consist of videos from various domains. DSH consists of six domains, namely, dog show, gymnastics, parkour, skating, skiing, and surfing. Each domain contains about 100 videos, and the total duration over all the six domains is around 1,430 minutes. TVSum consists of 50 videos, totaling around 210 minutes, from ten domains, namely, beekeeping (BK), bike tricks (BT), dog show (DS), flash mob (FM), grooming animal (GA), making sandwich (MS), parade (PR), parkour (PK), vehicle tire (VT), and vehicle unstuck (VU). However, neither of these datasets provide any user preferences. Therefore, to test our method on these datasets, we use the testing setup with no preferred highlight clips for any user (Sec.~4.3). We show the results on the DSH dataset in Table~\ref{tab:map_dsh}, and on the TVSum dataset in Table~\ref{tab:map_tvsum}. The results of all the methods we compare with are taken from Bhattacharya et al.~\cite{highlightme}. We observe that our method has the second-best mean average precision of all the methods over all the domains. Our method is also at par with the human-centric approach of HighlightMe~\cite{highlightme} for most of the domains, being only significantly outperformed in domains that have an abundance of both face and pose modalities and no other detected objects in the highlighted segments, \textit{e.g.}, flash mob (FM) and parade (PR) in TVSum. The precision of our method also suffers in the surfing domain in DSH and the bike tricks (BT) domain in TVSum where our method assigns highlight labels to many objects and human interactions not related to the ground-truth highlights (such as prepping surfboards, people interviewing, cars on the street). As part of our future work, for such scenarios, we can incorporate more human-centric modalities into our pipeline or provide domain-specific videos as the preferred clips to improve performance.

\begin{table*}[t]
    \centering
    \caption{Mean average precision on the DSH dataset~\cite{lsvm_dsh}. \textbf{Bold} indicates best, \underline{underline} indicates second-best.}
    \label{tab:map_dsh}
    \begin{tabular}{lcccccc}
        \toprule
        Domain & RRAE \cite{rrae} & Video2 GIF \cite{video2gif} & LSVM \cite{lsvm_dsh} & Less is More \cite{less_is_more} & HighlightMe~\cite{highlightme} & Ours \\
        \midrule
        dog show & 0.49 & 0.31 & 0.60 & 0.58 & \textbf{0.63} & \underline{0.60} \\
        gymnastics & 0.35 & 0.34 & 0.41 & \underline{0.44} & \textbf{0.73} & \textbf{0.73} \\
        parkour & 0.50 & 0.54 & 0.61 & 0.67 & \textbf{0.72} & \underline{0.71} \\
        skating & 0.25 & 0.55 & 0.62 & 0.58 & \textbf{0.64} & \underline{0.62} \\
        skiing & 0.22 & 0.33 & 0.36 & 0.49 & \underline{0.52} & \textbf{0.61} \\
        surfing & 0.49 & 0.54 & 0.61 & \textbf{0.65} & \underline{0.62} & 0.58 \\
        \midrule
        Mean & 0.38 & 0.46 & 0.54 & 0.57 & \textbf{0.64} & \underline{0.62} \\
        \bottomrule
    \end{tabular}
\end{table*}

\begin{table*}[t]
    \centering
    \caption{Mean average precision on the TVSum dataset~\cite{tvsum}. \textbf{Bold} indicates best, \underline{underline} indicates second-best.}
    \label{tab:map_tvsum}
    \begin{tabular}{lccccccc}
        \toprule
        Domain & MBF \cite{mbf} & KVS \cite{category_specific_summ} & CVS \cite{collab_summ} & Adv-LSTM \cite{unsup_video_summ} & Less is More \cite{less_is_more} & HighlightMe~\cite{highlightme} & Ours \\
        \midrule
        BK & 0.31 & 0.34 & 0.33 & 0.42 & \textbf{0.66} & 0.57 & \underline{0.60} \\
        BT & 0.37 & 0.42 & 0.40 & 0.48 & 0.69 & \textbf{0.93} & \underline{0.85} \\
        DS & 0.36 & 0.39 & 0.38 & 0.47 & \textbf{0.63} & 0.60 & \underline{0.62} \\
        FM & 0.37 & 0.40 & 0.37 & 0.46 & 0.43 & \textbf{0.88} & \underline{0.77} \\
        GA & 0.33 & 0.40 & 0.38 & 0.48 & 0.61 & \underline{0.50} & \textbf{0.63} \\
        MS & 0.41 & 0.42 & 0.40 & 0.49 & \underline{0.54} & 0.50 & \textbf{0.55} \\
        PR & 0.33 & 0.40 & 0.38 & 0.47 & 0.53 & \textbf{0.84} & \underline{0.70} \\
        PK & 0.32 & 0.38 & 0.35 & 0.46 & 0.60 & \textbf{0.76} & \underline{0.68} \\
        VT & 0.30 & 0.35 & 0.33 & 0.42 & 0.56 & \textbf{0.65} & \underline{0.61} \\
        VU & 0.36 & 0.44 & 0.41 & 0.47 & 0.50 & \textbf{0.77} & \underline{0.75} \\
        \midrule
        Mean & 0.35 & 0.40 & 0.37 & 0.46 & 0.58 & \textbf{0.70} & \underline{0.68} \\
        \bottomrule
    \end{tabular}
\end{table*}

\end{document}